\documentclass{article}

\pdfoutput=1

     \PassOptionsToPackage{numbers, compress}{natbib}

     \usepackage[preprint]{neurips_2020}

\usepackage[utf8]{inputenc} 
\usepackage[T1]{fontenc}    
\usepackage{hyperref}       
\usepackage{url}            
\usepackage{booktabs}       
\usepackage{amsfonts}       
\usepackage{nicefrac}       
\usepackage{microtype}      
\usepackage{microtype}
\usepackage{graphicx}
\usepackage{subfloat}
\usepackage{booktabs} 
\usepackage{subfig}
\usepackage{multirow}
\usepackage{subfig}
\usepackage{listings}
\usepackage{multicol}

\usepackage{amsmath,amsfonts,bm}

\def\figref#1{figure~\ref{#1}}
\def\Figref#1{Figure~\ref{#1}}

\def\secref#1{section~\ref{#1}}

\def\eqref#1{equation~\ref{#1}}

\def\tabref#1{table~\ref{#1}}

\def\1{\bm{1}}

\def\rx{{\textnormal{x}}}

\def\rvx{{\mathbf{x}}}
\def\rvy{{\mathbf{y}}}
\def\rvz{{\mathbf{z}}}

\def\ve{{\bm{e}}}

\def\vx{{\bm{x}}}

\def\vz{{\bm{z}}}

\DeclareMathAlphabet{\mathsfit}{\encodingdefault}{\sfdefault}{m}{sl}
\SetMathAlphabet{\mathsfit}{bold}{\encodingdefault}{\sfdefault}{bx}{n}

\def\sR{{\mathbb{R}}}

\newcommand{\E}{\mathbb{E}}

\newcommand{\KL}{D_{\mathrm{KL}}}

\usepackage{xcolor,colortbl}
\definecolor{green}{rgb}{0.91764706, 0.94117647, 0.86666667}
\definecolor{red}{rgb}{0.94509804, 0.85490196, 0.85490196}
\newcommand{\green}[1]{\cellcolor{green}#1}  
\newcommand{\red}[1]{\cellcolor{red}#1}  

\title{Anomaly localization by modeling perceptual features}

\author{David Dehaene\thanks{Equal contributions.}, \, Pierre Eline\footnotemark[1] \\
AnotherBrain, Paris, France\\
\texttt{\{david, pierre\}@anotherbrain.ai} \\
}

\begin{document}

\maketitle

\begin{abstract}
Although unsupervised generative modeling of an image dataset using a Variational AutoEncoder (VAE) has been used to detect anomalous images, or anomalous regions in images, recent works have shown that this method often identifies images or regions that do not concur with human perception, even questioning the usability of generative models for robust anomaly detection.
Here, we argue that those issues can emerge from having a simplistic model of the anomaly distribution and we propose a new VAE-based model expressing a more complex anomaly model that is also closer to human perception.
This Feature-Augmented VAE is trained by not only reconstructing the input image in pixel space, but also in several different feature spaces, which are computed by a convolutional neural network trained beforehand on a large image dataset.
It achieves clear improvement over state-of-the-art methods on the MVTec anomaly detection and localization datasets.
\end{abstract}
\section{Introduction}

The use of large databases of labeled data has been extensively leveraged in the past decade to automate industrial tasks such as digitalization of handwritten letters or identification of cancerous tissue.
In the domain of quality control, in which a given image is either identified as normal or anomalous, labels would need to cover the entirety of the possible anomalies range (stain, crack, ...) for such an approach to be used on a production line. 
When anomalies are rare, the constitution of such an exhaustive dataset can be time-consuming.
The unsupervised anomaly detection setting is a two-class classification task in which the training set is only comprised of non-defective samples, from the \textit{normal} class, whereas samples from the second, \textit{anomalous} class are only given at test time.
For quality control, this task can also be extended to a pixel-wise segmentation of anomalous regions on otherwise normal samples, which we will call anomaly localization.

\begin{figure}[h!]
\subfloat[Subfigure 0 list of figures text][]{
\includegraphics[width=0.22\textwidth]{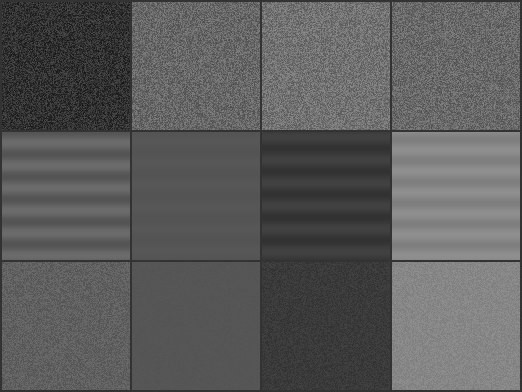}}
\hskip1em\relax
\subfloat[Subfigure 1 list of figures text][]{
\includegraphics[width=0.22\textwidth]{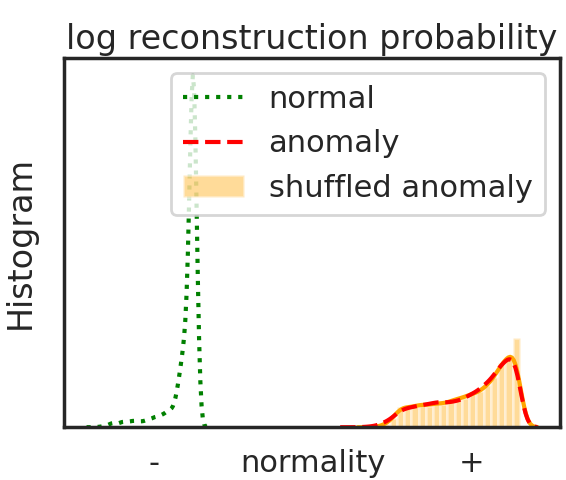}
\label{fig:histograms1}}
\hskip1em\relax
\subfloat[Subfigure 2 list of figures text][]{
\includegraphics[width=0.22\textwidth]{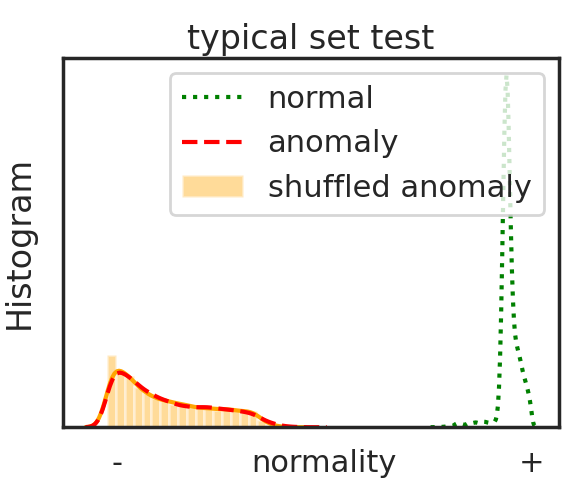}
\label{fig:histograms2}}
\hskip1em\relax
\subfloat[Subfigure 3 list of figures text][]{
\includegraphics[width=0.22\textwidth]{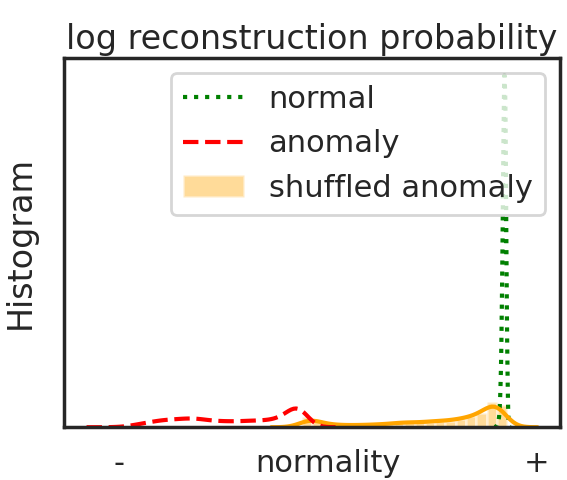}
\label{fig:histograms3}}
\caption{(a) Line by line, respectively: samples from \textit{normal} dataset, samples from \textit{anomaly} dataset, samples from \textit{shuffled anomaly} dataset obtained by shuffling \textit{anomaly} pixel wise.
(b) The classic VAE approach, despite a model learned to perfectly represent the normal distribution, ranks the anomalous samples as more likely to be drawn from the normal distribution than the normal samples. (c) On a classic VAE, the typicality test rejects the anomalous samples in accordance to human perception, but the same test would reject the shuffled anomalous samples, whereas a human would probably not. (d) Our model's log probability separates classes in accordance to human perception.}
\label{fig:histograms}
\end{figure}

Variational AutoEncoders (VAE, \citet{kingma2013auto}) are especially good models for this task, since they approximately model the probability density of the distribution of their training dataset.
Anomaly detection can then be seen as thresholding the approximated sample probability: a sample with low probability under the evaluated normal data distribution is seen as an anomaly.
Moreover, they are also useful for anomaly localization. 
Indeed, autoencoders learn to reconstruct training samples after compressing them to a low dimensional space, training encoder (compressive) and decoder (decompressive) models jointly with a distance-based metric between an original sample and its reconstruction.
When used on an anomalous sample, a VAE will reconstruct correctly normal regions of the image, whereas anomalous regions may contain never-before-seen features that the decoder will struggle to imitate.
A number of works thus compared the input sample with its reconstruction pixel-wise to localize anomalies in an image \citep{Bergmann2019, An2015, Baur2018, Matsubara2018}.

Nevertheless, on the one hand \citet{nalisnick2018deep} questioned the feasibility of detecting anomalies using generative models, showing that a VAE trained on the CIFAR10 dataset \citep{cifar10} assigned higher probability values to samples from the fairly different SVHN dataset \citep{svhn} than to test samples from CIFAR10.
On the other hand, \citet{Bergmann2018} identified issues of accordance with human perception of anomalies with a pixel-wise reconstruction error, and used the Structural SIMilarity (SSIM, \citet{Wang2004}) to identify larger zones of perceptual anomalies.

In this paper, we argue that these two problems are linked and we contribute two new limitations of the standard reconstruction based framework:
\begin{itemize}
    \item Training on normal data only can make the model "blind" to anomalous features present in defective data whereas humans are very sensitive to some of those features.
    \item High likelihood of a sample under the normal data distribution does not necessarily mean high probability of being normal. A comparison with the likelihood of the sample under the (unknown) anomalous data distribution is required.
\end{itemize}
We illustrate those limitations with a toy dataset.
By leveraging pre-trained networks on large image datasets, we contribute a new reconstruction-based model on both pixel and perceptual features spaces. 
We demonstrate that it outperforms state-of-the-art results on all subsets of the MVTec AD dataset \citep{Bergmann2019}.


\section{Related work}

\subsection{Anomaly localization using generative models}
Generative models are trained to approximate the distribution of the training set by learning to generate samples from that distribution.
Generative Adversarial Networks (GAN, \citet{goodfellow2014generative}) do so by jointly training a generator network to generate credible samples from noise, and a discriminator network to classify the samples as either generated or real.
The two models implicitly define the training data manifold as the union of samples that the generator can generate and samples that the discriminator labels as real.
\citet{SCHLEGL201930, schlegl2017unsupervised} use GAN for anomaly detection and localization by searching for the closest neighbor of a given sample on the manifold.
They compare the sample with this normal closest neighbor to identify the presence and location of an anomaly.

VAE \citep{kingma2013auto} present a more direct estimation of the normal distribution's density as a lower bound.
The goal is to approximate the true distribution $q_n(\rvx)$, $\rvx \in  \sR^d$, with $d$ the dimension of the input space, using the generative model $p_\theta(\rvx, \rvz) = p_\theta(\rvx|\rvz) \, p(\rvz)$, with $\rvz \in \sR^l$, $l < d$, and conditional distributions output by the decoder $p_\theta(\rvx|\rvz)$ ($\theta$ representing the decoder model's parameters). An importance sampling distribution $q_\phi(\rvz|\rvx)$ output by the encoder model ($\phi$ representing the encoder model's parameters) is introduced to reduce the estimator's variance. Jensen's inequality gives the Evidence Lower Bound Objective (ELBO) with which the VAE is trained:
\begin{equation}
\E_{q_n(\rvx)} \log p_\theta (\rvx) \geq  \E_{q_n(\rvx)} [ \; \underbrace{\E_{q_\phi(\rvz|\rvx)} \log p_\theta(\rvx|\rvz)}_{\text{reconstruction term}} - \underbrace{\KL(q_\phi(\rvz|\rvx)\Vert p(\rvz))  }_{\text{regularization term}} \;]
\label{eq:vae2}
\end{equation}

We will note $\mu_{\theta}(\rvz)$ the mean of the decoded distribution $p_\theta(\rvx|\rvz)$ and $\vz_{\phi}(\rvx)$ the mean of the encoded distribution $q_\phi(\rvz|\rvx)$.
The VAE \textit{reconstruction} of a given sample $\rvx$ is defined as $r_{\theta, \phi}(\rvx) = \mu_{\theta}(\vz_{\phi}(\rvx))$.
As with GAN, VAE are mainly used for anomaly detection and localization by comparing a sample and its reconstruction \citep{Bergmann2018, Bergmann2019, Baur2018, Matsubara2018}.
\citet{Bergmann2019} show that VAE achieves better results than GAN on a variety of anomaly localization datasets. 
Indeed, it is a known GAN flaw \citep{goodfellow2016nips} that even though they typically generate images of better quality than VAE, they have a tendency to only be able to generate a subset of the training distribution.
This is especially problematic for anomaly detection, in which a whole subset of the normal distribution could be flagged as anomalous.

\citet{An2015} proposed the ELBO as an anomaly score to classify samples. \citet{Matsubara2018} used only the reconstruction term of the ELBO to localize anomalies in images.
\citet{Baur2018} used several variants of VAE and GAN like architectures to localize the differences between a sample and its reconstruction.
\citet{Bergmann2018} replaced the pixel-wise L2 reconstruction error with the SSIM \citep{Wang2004}, then \citet{Bergmann2019} showed that autoencoder-based models were the best architectures for anomaly localization on the MVTec AD dataset, an anomaly localization dataset that they submitted, but also that depending on the dataset, standard autoencoders could achieve better performance than SSIM autoencoders.
\citet{Dehaene2020} showed that a local image anomaly could deteriorate the VAE reconstruction of a whole image, reducing the effectiveness of reconstruction-based anomaly localization, and proposed a gradient-based method similar to AnoGAN to refine anomaly segmentations given by autoencoder-based models.

\subsection{Anomaly detection using transfer learning}

In anomaly detection, the training dataset is only comprised of normal samples.
Some features that are present during test time were never seen by the model during training but may be crucial for normal/abnormal classification.
The transfer of discriminative features from pre-trained models, while common in computer vision, has only recently been used for anomaly localization.

\citet{Napoletano2018} used the distance to the nearest neighbor in a set of trained normal prototypes in a pre-trained feature space as an anomaly score.
\citet{nazare2018pretrained} modeled these feature distributions using K-Means.
They both used high-level features in their pre-trained models, so that to use their models for anomaly localization, one has to treat images as overlapping patches.
\citet{Sabokrou_2018} enabled fast localization by using the convolutional features of a pre-trained network and modeled them as one-dimensional normal distributions.
\citet{Andrews2016TransferRF, Burlina_2019_CVPR} used the features extracted from pre-trained models to train a One-Class Support Vector Machine (OC SVM) to achieve better results than a pixel-wise OC SVM for anomaly detection.

Very recently, \citet{bergmann2019uninformed} compared those previous models as well as autoencoder reconstruction-based methods.
They also proposed a method based on an ensemble of regressors, trained to imitate a pre-trained CNN (the teacher) on the training dataset.
The regressors are supposed to reproduce the teacher's features well on the training dataset, but to diverge from the teacher, as well as from the other regressors, on anomalous patches, so that an anomaly map can be derived from the ensemble.

In this paper, we merge the transfer learning approach with the anomaly localization framework given by the VAE.

\section{Implicit biases in previous approaches}
\subsection{The example distribution}
\label{sec:toy}


To understand the problem we are trying to tackle, in this section we will study a toy dataset of grayscale images.
Formal equations defining the normal and anomalous distributions are in Appendix 1, and samples from the three distributions are shown in \figref{fig:histograms} (a).
Normal images are composed of pixel-wise white noise on a random monochrome background.
This is a very simple latent variable model that a VAE can learn to approximate perfectly, so that $\KL ( q_n(\rvx) \Vert p_\theta(\rvx) ) = 0$ (Proof in Appendix 2).
Anomalous images are composed of a background of horizontal stripes, with added white noise of an amplitude smaller than the noise amplitude of the normal data distribution.
Finally we introduce a third distribution, which we call the shuffled anomaly distribution.
To sample from this distribution, we first sample from the anomaly distribution, then we randomly shuffle the pixels in the sample.

We use a VAE with the decoder variance $\gamma$ trained as a global parameter: $p_{\theta}(\rvx|\rvz) \equiv \mathcal{N}(\mu_{\theta}(\rvz), \gamma^2)$ as described in \citet{Dai2019}.
Even though it is obvious for the human reader which samples are anomalies, we can see in \figref{fig:histograms} (b) that in this setting the VAE affects higher likelihoods to anomalous samples than to normal samples.
This is even more surprising when we note that the normal distribution is simple enough for the VAE to model perfectly.
The result of \figref{fig:histograms} (b) is in fact inherent to the two chosen distributions.

\subsection{Typical set score}

In this subsection we show that previous approaches solely based on statistical tests are not sufficient if they do not align with human perception.
\citet{nalisnick2019detecting, choi2018waic} studied a similar thought experiment as in \secref{sec:toy}, with constant $\mu$ and $f = 0$, and proposed a typicality test based on the notion of typical sets as a solution.

Given a set ${\rvx_1..., \rvx_N}$ of samples to classify,
$- |\frac{1}{N} \log p_{\theta}(\rvx_1,...,\rvx_N) + H[p_{\theta}(\rvx)]|$ is proposed as an anomaly score for the set.
The intuition behind the use of typical sets is to accumulate evidence over several independent realizations of a random process to determine if the random process follows a given distribution.
The drawback of this accumulation is that the score is no more associated with a single sample, but to the whole set.
To use this idea, while still getting a single-image analysis, we will consider the realizations $\epsilon_k = [ \,\rvx_k - \mu_{\theta}(\rvz)_k \,]$ over all pixels $k$, which are i.i.d. with distribution $\mathcal{N}(0, \gamma^2)$, to accumulate evidence across all pixels and generate an image-wide diagnostic:
\begin{equation}
\text{Typical score} = - | \frac{1}{d} \sum_{k=1}^d \log p_{\theta}([ \,\rvx_k - \mu_{\theta}(\rvz)_k \,]|\rvz) + H[\mathcal{N}(0, \gamma^2)] |
\end{equation}
We can see in \figref{fig:histograms} (c) that the test indeed reorders the normal and anomaly distribution in accordance to human perception.
Nevertheless, since the typicality test is invariant to a shuffle in pixel ordering, it cannot separate the shuffled anomaly distribution and the anomaly distribution.
This can be problematic because for a human assessor, samples from the shuffled anomaly distribution are similar to normal samples.
This statistical test, while theoretically sound to reject atypical events, is not able to yield the perceptive ordering of the three distributions.

\subsection{Implicit models of anomalies}

How can we have a perfect approximation of the normal distribution and reject different abnormal samples than a human on such a simple case ?  
What seems to be a paradox here should not really be one: it is not generally possible to discriminate normal from anomalous data using only the normal distribution's density $q_n(\rvx) = q(\rvx|n)$, since we need to compare it to the abnormal distribution's density $q_a(\rvx) = q(\rvx|a)$.
The optimal normal/abnormal classifier $C_a^*(\rvx)$ uses both distributions:
\begin{equation}
C_a^*(\rvx) = \frac{q(a|\rvx)}{q(n|\rvx)} = \frac{q_a(\rvx) q(a)}{q_n(\rvx) q(n)}
\end{equation}
with $q(a) = 1 - q(n) = \int_{\rvx} q(a|\rvx) $.
Here we will explicit the anomaly models implicitly used by classic approaches in order to introduce our anomaly model.

The standard VAE approach to anomaly localization is only based on the quality of the pixel-wise reconstruction: 
one classically defines a pixel-wise anomaly classifier $C_{a_k}(\rvx)$ using a threshold T so that
\begin{equation}
\forall k, \;\; (C_{a_k}(\rvx) > 0.5) \iff (p_{\theta}(\rx_k|\vz_{\phi}(\rvx)) < T)
\label{eq:classic_classifier}
\end{equation}
This can explicitly be obtained (in Appendix 3) by seeing the tested image as the composition of a sample from a more restrictive normal distribution $\mu_{\theta}(\rvz)$ and white pixel noise $\ve$ from (for example) a Gaussian distribution.
This restrictive normal distribution is obtained from deterministically decoding latent variables and assigning them the mean of the decoded distribution. This gives us an anomalous sample $\rvx_a$ as:
\begin{equation}
\begin{split}
\rvx_a &= \mu_{\theta}(\rvz) + \ve \\
\rvz \sim p(\rvz), \;\;\;\; &\ve \sim p_a(\ve) \equiv \mathcal{N}(0, 
\sigma_a^2)
\end{split}
\label{eq:classic_anomaly_model}
\end{equation}
with $\sigma_a$ a constant.

This models anomalies as independent noise over the pixels, which is not a very informative prior for image analysis.

\section{Proposed method}
\subsection{Transferable features for anomaly modeling}
We want to modify the anomaly model from \eqref{eq:classic_anomaly_model} to model more complex and perceptual anomalies.
This may be done by transforming the anomalies in a feature space in which we could model our test images as 
\begin{equation}
\begin{split}
\rvx_a &= \vx_n + \ve \\
f(\rvx_a) &= f(\vx_n) + \ve_f \\
\vx_n \sim q_n(\rx_n), &\;\;\; \ve_f \sim \mathcal{N}(0, \sigma_{fa})
\end{split}
\label{eq:our_anomaly_model}
\end{equation}

with $f$ some transformation of the pixels.

It is now the features of the image that are disturbed with white noise in an abstract feature space.
We nevertheless have to find the right feature space $f(\sR^d)$ so that this model is useful for the detection of anomalies.
We characterize its usefulness with the following properties:
\begin{itemize}
\item It should be local to enable a precise localization of anomalies.
\item It should be robust to a large variety of possible anomalies, of different sizes and complexities.
\item It should be as close as possible to the human perception, i.e. enable the detection and localization of the same kind of anomalies as human assessors.
\end{itemize}
Concerning the last point, convolutional neural networks trained beforehand on large image datasets are often used to define similarity metrics that are close to human perception \citep{Zhang_2018, Hou_2017}, so their feature maps are great answers to that problem.
To make the localization robust and precise, we will use $L$ feature maps, taken at different layers of a pre-trained CNN, such as VGG16 \citep{simonyan2014deep}, that we will call the feature extractor.
Going up through the layers, we will trade precision in the localization for a more complex structure of the anomalies.

Now that we have selected our $f$ transforms, we need to have a model of the normal distribution in these feature spaces, an approximation of the $f(\vx_n)$ term in \eqref{eq:our_anomaly_model}. 
This can be done as in \eqref{eq:classic_anomaly_model}, using VAEs.

\subsection{Feature Augmented VAE}

For each feature extractor layer $f_i, i \in [1..L]$ we use a VAE to approximate its distribution on the normal dataset:
\begin{equation}
\begin{split}
\E_{q_n(\rvx)} \log p_{\theta_0}(\rvx) &\geq ELBO_{\phi_0, \theta_0}(\rvx) \\
\forall i \in [1..L], \E_{q_n(\rvx)} \log p_{\theta_i}(\rvy_i) &\geq ELBO_{\phi_i, \theta_i}(\rvy_i) \text{ with } \rvy_i = f_i(\rvx)
\end{split}
\end{equation}
Each channel of each feature space is given an independent decoder variance parameter $\gamma_{i, c}^2$, which is trained globally and is independent from the latent space.

\begin{figure}[h!]
\subfloat[Architecture of the model]{
\includegraphics[width=0.45\textwidth]{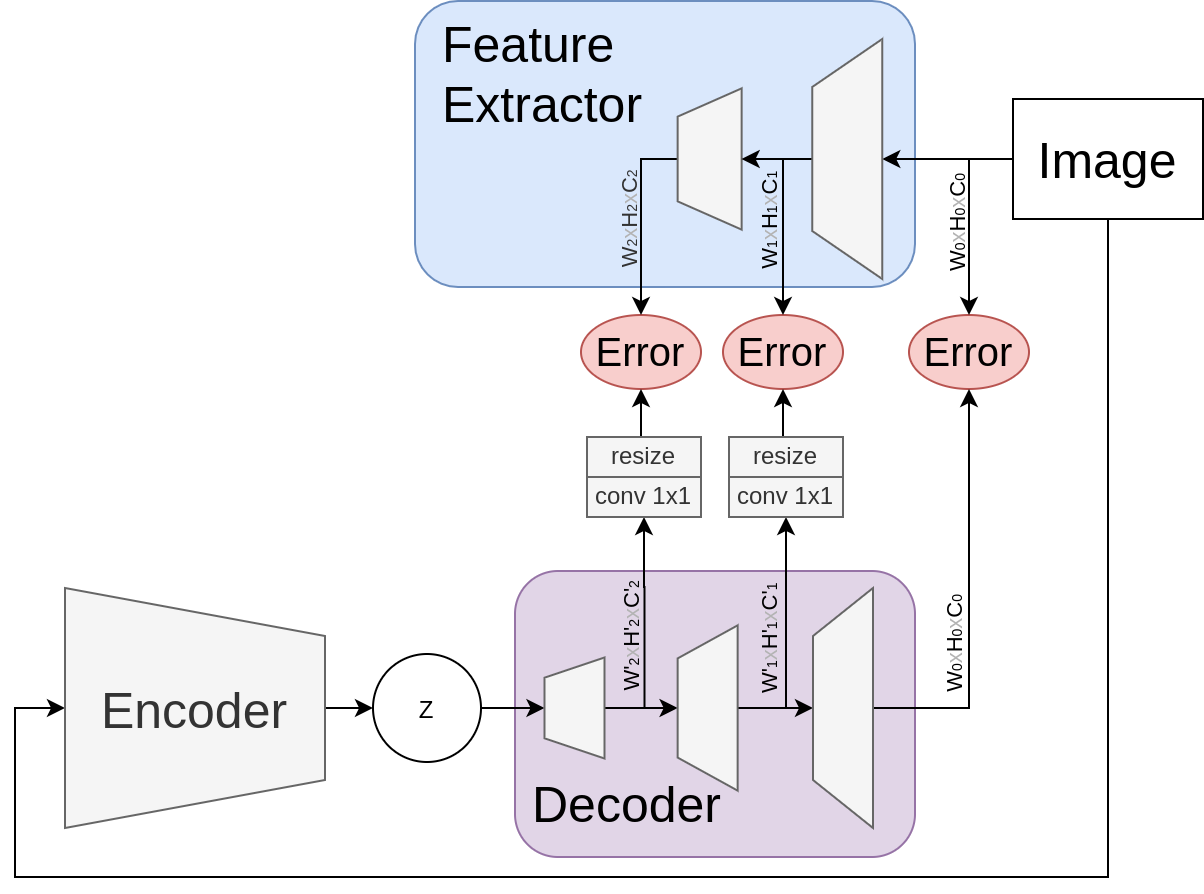}
\label{fig:architecture}}
\hskip0.05\textwidth \relax
\subfloat[Anomaly map computations]{\includegraphics[width=0.45\textwidth]{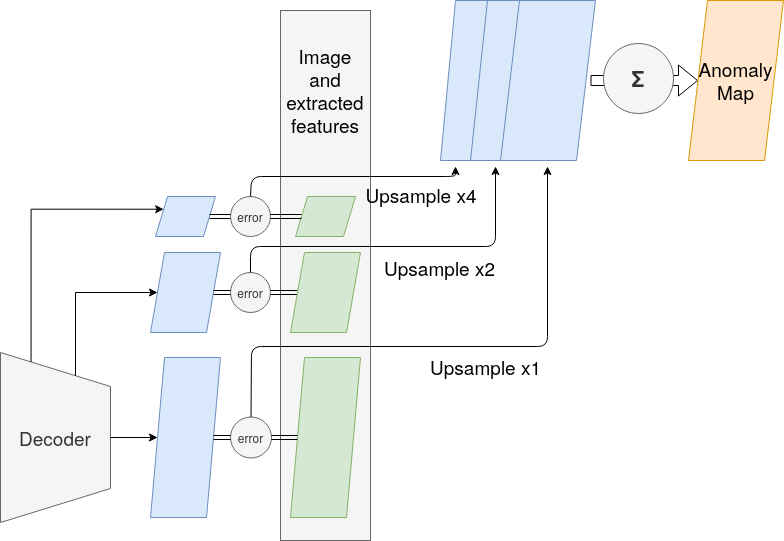}
\label{fig:fusion}}
\caption{Illustrations of the model (a) and the anomaly map computations (b) for $L$ = 2}
\end{figure}

To reduce the number of parameters used by the model, we propose to use only one encoder and decoder for all the features and pixels to reconstruct.
The features of higher level in the feature extractor will be decoded in early layers of the decoder, whereas its output remains a distribution over the pixels.
A 1x1 kernel convolution followed by a spatial bilinear interpolation are used if the number of channels and spatial resolution in the decoder and feature extractor feature map do not match.
The architecture of the model is drawn in \figref{fig:architecture}.
The time and space complexity of this model is roughly the same as the complexity of the underlying VAE model added to the complexity of the feature extractor.
The latter is not trained: its weights are frozen during the VAE's training.

For training as well as for anomaly detection, the merging of the different layers' errors is done by spatially upsampling the error maps of each layers to the original image resolution, denoted by a hat: $\widehat{\rvy_i} \in \sR^d$.
The $L$ error maps are then summed over the layers dimension to get a single error map the size of the original image.
This process is illustrated in \figref{fig:fusion}.
The full loss function for the Feature Augmented VAE (FAVAE) is:
\begin{equation}
\begin{split}
\mathcal{L} = -\sum_{i=0}^{L} \, \left[ \E_{q_\phi(\rvz|\rvx)}\widehat{\log p_{\theta}(\rvy_i|\rvz)} \right]
+ \KL(q_\phi(\rvz|\rvx)\Vert p(\rvz))
\end{split}
\end{equation}
with $\rvy_0 = \rvx$



The anomaly score for a pixel is computed by calculating a log probability map for each layer (summing only on channels), then by upsampling each map to the size of the original image, and finally by summing over the layers, as in \figref{fig:fusion}.

We define the anomaly score for the whole image as the sum of this map over every pixel and channel.

To conclude this section, \figref{fig:histograms}d shows that using FAVAE, a form of perceptual anomaly detection has been reached on the toy dataset: the three datasets are ranked in a perceptually conform order.

\section{Experiments}

\subsection{MVTec Anomaly Detection dataset}

\definecolor{gray}{rgb}{0.9, 0.9, 0.9}
\newcommand{\gray}[1]{\cellcolor{gray}#1}  
\begin{table*}[h]
\caption{MVTec results for FAVAE, Vanilla VAE and state of the art algorithms. \textit{Ref} are the best results from \citet{Dehaene2020}. For a given dataset and metric, green cell indicates that FAVAE performs better than vanilla VAE (on average) and  best result among all models is written in bold. Each experiment has been run 5 times. We report mean and standard deviation for each experiment.}
\newcommand{\doublecell}[1]{\multirow{2}{1.3cm}{\centering \textbf{#1}}}
\begin{center}

 \begin{tabular}{|c|c|c|c||c|c|}
    \hline
  	\multirow{2}{1cm}{\textbf{Dataset}} &
    \multicolumn{3}{c||}{\textbf{Pixel AUROC}} &
    \multicolumn{2}{c|}{\textbf{Image AUROC}} \\ \cline{2-6}
    & \textbf{Ref} & \textbf{FAVAE} & \textbf{Vanilla VAE}  
    & \textbf{Vanilla VAE} & \textbf{FAVAE} \\ \hline

 carpet &   0.774 \tiny - & \green \textbf{ 0.960} \tiny $\pm$ 0.004 &   0.620 \tiny $\pm$ 0.006 &   0.515 \tiny $\pm$ 0.011 & \green \textbf{ 0.671} \tiny $\pm$ 0.010\\ \hline 
 grid &   0.981 \tiny - & \green \textbf{ 0.993} \tiny $\pm$ 0.000 &   0.856 \tiny $\pm$ 0.003 &   0.904 \tiny $\pm$ 0.037 & \green \textbf{ 0.970} \tiny $\pm$ 0.004\\ \hline 
 leather &   0.925 \tiny - & \green \textbf{ 0.981} \tiny $\pm$ 0.001 &   0.835 \tiny $\pm$ 0.015 &   0.619 \tiny $\pm$ 0.043 & \green \textbf{ 0.675} \tiny $\pm$ 0.013\\ \hline 
 tile &   0.654 \tiny - & \green \textbf{ 0.714} \tiny $\pm$ 0.001 &   0.520 \tiny $\pm$ 0.005 &   0.512 \tiny $\pm$ 0.030 & \green \textbf{ 0.805} \tiny $\pm$ 0.008\\ \hline 
 wood &   0.838 \tiny - & \green \textbf{ 0.899} \tiny $\pm$ 0.002 &   0.699 \tiny $\pm$ 0.002 &   0.875 \tiny $\pm$ 0.009 & \green \textbf{ 0.948} \tiny $\pm$ 0.002\\ \hline 
 bottle &   0.951 \tiny - & \green \textbf{ 0.963} \tiny $\pm$ 0.000 &   0.894 \tiny $\pm$ 0.002 &   0.985 \tiny $\pm$ 0.002 & \green \textbf{ 0.999} \tiny $\pm$ 0.001\\ \hline 
 cable &   0.910 \tiny - & \green \textbf{ 0.969} \tiny $\pm$ 0.001 &   0.816 \tiny $\pm$ 0.014 &   0.805 \tiny $\pm$ 0.008 & \green \textbf{ 0.950} \tiny $\pm$ 0.003\\ \hline 
 capsule &   0.952 \tiny - & \green \textbf{ 0.976} \tiny $\pm$ 0.001 &   0.907 \tiny $\pm$ 0.010 &   0.694 \tiny $\pm$ 0.038 & \green \textbf{ 0.804} \tiny $\pm$ 0.008\\ \hline 
 hazelnut &  \textbf{ 0.988} \tiny - & \green  0.987 \tiny $\pm$ 0.000 &   0.951 \tiny $\pm$ 0.002 &   0.922 \tiny $\pm$ 0.016 & \green \textbf{ 0.993} \tiny $\pm$ 0.001\\ \hline 
 metal nut &   0.920 \tiny - & \green \textbf{ 0.966} \tiny $\pm$ 0.001 &   0.861 \tiny $\pm$ 0.009 &   0.676 \tiny $\pm$ 0.010 & \green \textbf{ 0.852} \tiny $\pm$ 0.004\\ \hline 
 pill &   0.935 \tiny - & \green \textbf{ 0.953} \tiny $\pm$ 0.001 &   0.879 \tiny $\pm$ 0.003 &   0.808 \tiny $\pm$ 0.008 & \green \textbf{ 0.821} \tiny $\pm$ 0.009\\ \hline 
 screw &   0.983 \tiny - & \green \textbf{ 0.993} \tiny $\pm$ 0.003 &   0.928 \tiny $\pm$ 0.028 &   0.654 \tiny $\pm$ 0.111 & \green \textbf{ 0.837} \tiny $\pm$ 0.101\\ \hline 
 toothbrush &   0.985 \tiny - & \green \textbf{ 0.987} \tiny $\pm$ 0.000 &   0.953 \tiny $\pm$ 0.002 &  \textbf{ 0.987} \tiny $\pm$ 0.004 & \red  0.958 \tiny $\pm$ 0.006\\ \hline 
 transistor &   0.934 \tiny - & \green \textbf{ 0.984} \tiny $\pm$ 0.001 &   0.851 \tiny $\pm$ 0.006 &   0.871 \tiny $\pm$ 0.007 & \green \textbf{ 0.932} \tiny $\pm$ 0.003\\ \hline 
 zipper &   0.889 \tiny - & \green \textbf{ 0.968} \tiny $\pm$ 0.002 &   0.775 \tiny $\pm$ 0.011 &   0.797 \tiny $\pm$ 0.138 & \green \textbf{ 0.972} \tiny $\pm$ 0.002\\ \hline 
\gray mean & \gray  0.908 \tiny $\pm$ 0.088 & \green \textbf{ 0.953} \tiny $\pm$ 0.068 & \gray  0.823 \tiny $\pm$ 0.120 & \gray  0.775 \tiny $\pm$ 0.159 & \green \textbf{ 0.879} \tiny $\pm$ 0.108\\ \hline 

\end{tabular}

\end{center}
\label{table:backbone}

\end{table*}

In order to evaluate the proposed method for the task of anomaly segmentation, we perform experiments with the recently proposed MVTec AD dataset \citep{Bergmann2019}.
This collection of datasets comprises 5 categories of textures and 10 categories of objects in the context of industrial inspection, with 73 differents types of defects.
The quality of the samples are very close to what can be acquired in an industrial setting, and the diversity of the subsets allows to test a model on a large range of real life applications. 
For each category, we dispose of a training dataset containing normal samples and a test dataset containing normal and anomalous samples, with a labeled segmentation of the defects.

Our models are trained on normal training samples and tested on both normal and anomalous test samples to evaluate the anomaly segmentation performance.

As in \citet{Dehaene2020}, for the textures datasets, we first subsample the original dataset images to $512 \times 512$ and then crop random patches of size $128 \times 128$ which are used to train and test the different models.
For the object datasets, we subsample the original dataset images to $128 \times 128$ then we perform rotation and translation data augmentations. For all datasets we train on 10000 images.

Anomaly segmentation is then computed by reconstructing the anomalous image and its features and comparing those reconstructions with the original image augmented with its real features.
Given the ground truth segmentation from the test set, we compute the AUROC (Area Under the Receiver Operating Characteristics) at image (anomaly detection) and pixel level (anomaly localization). 
Note that an AUROC of 1 expresses the best possible classification in terms of normal and anomalous images/pixels.

\begin{figure}[h]
\subfloat[]{
\includegraphics[width=0.3\textwidth]{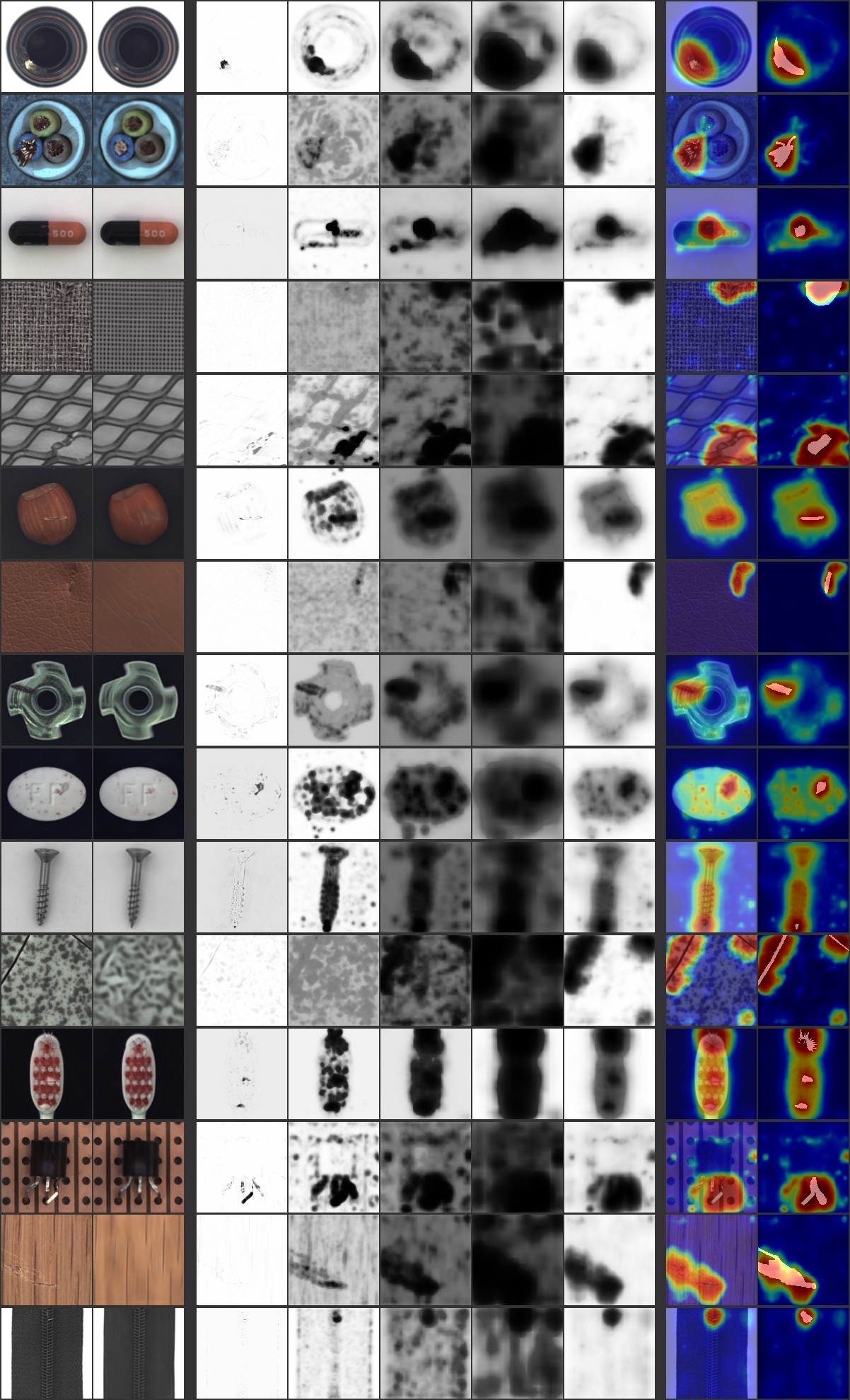}}
\hskip0.1\textwidth \relax
\subfloat[]{\newcounter{model}
\begin{tabular}[b]{l|l|c|c}
\multicolumn{1}{c|}{Model} &
\multicolumn{1}{c|}{Feature Extractor} &
\begin{tabular}[c]{@{}c@{}}Pixel\\AUROC\end{tabular} &
\begin{tabular}[c]{@{}c@{}}Image\\AUROC\end{tabular} \\ \hline
\begin{tabular}[c]{@{}l@{}}\includegraphics[width=0.07\textwidth]{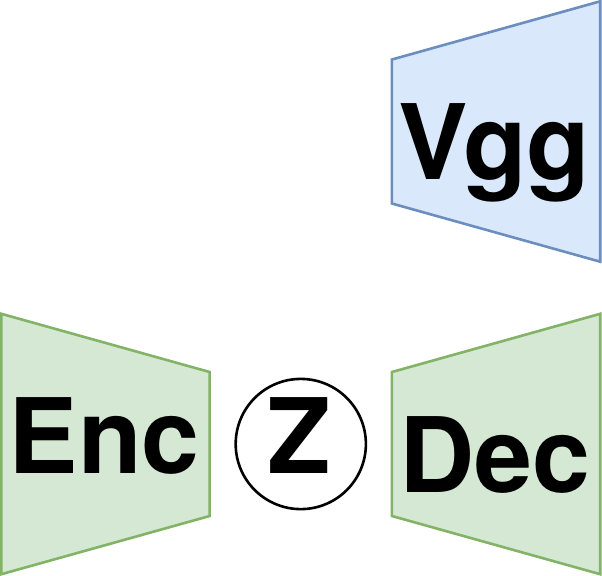}\end{tabular} & \refstepcounter{model}(\themodel)\label{model:m4} VGG16 & \textbf{0.953} & \textbf{0.883}\\ \hline
\begin{tabular}[c]{@{}l@{}}\includegraphics[width=0.07\textwidth]{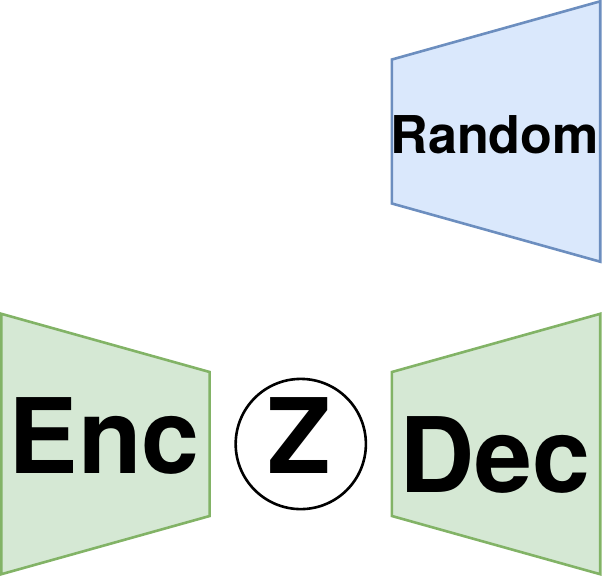}\end{tabular} & \refstepcounter{model}(\themodel)\label{model:m5} random & 0.899	& 0.764\\ \hline
\begin{tabular}[c]{@{}l@{}}\includegraphics[width=0.07\textwidth]{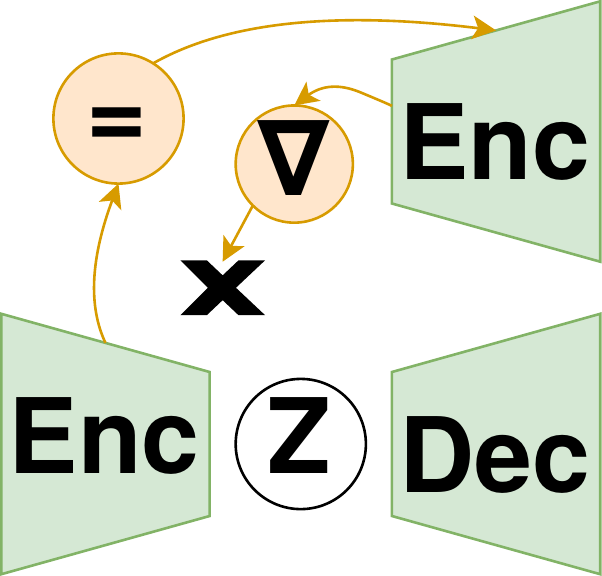}\end{tabular} &  \multicolumn{1}{l|}{\begin{tabular}[c]{@{}l@{}}\refstepcounter{model}(\themodel)\label{model:m3} encoder with \\ gradient stop \end{tabular}}  & 0.890 & 0.779\\ \hline
\begin{tabular}[c]{@{}l@{}}\includegraphics[width=0.07\textwidth]{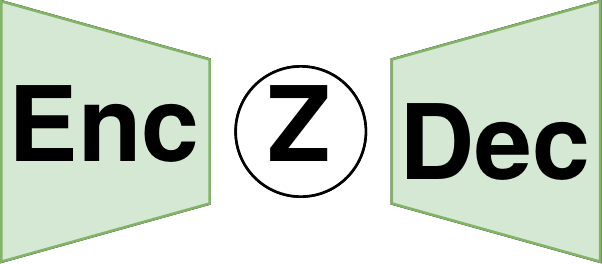}\end{tabular} & \refstepcounter{model}(\themodel)\label{model:m1} none & 0.824 &0.791\\ \hline
\begin{tabular}[c]{@{}l@{}}\includegraphics[width=0.07\textwidth]{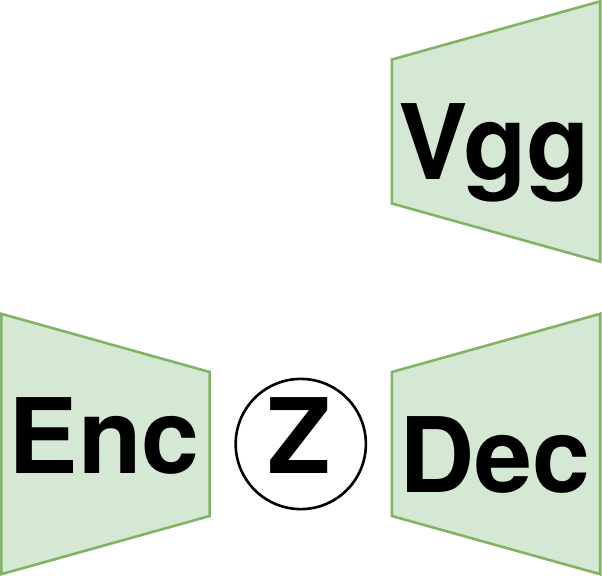}\end{tabular} & \multicolumn{1}{l|}{\begin{tabular}[c]{@{}l@{}}\refstepcounter{model}(\themodel)\label{model:m6} unfrozen \\ VGG16 \end{tabular}} & 0.762 & 0.686\\ \hline
\begin{tabular}[c]{@{}l@{}}\includegraphics[width=0.07\textwidth]{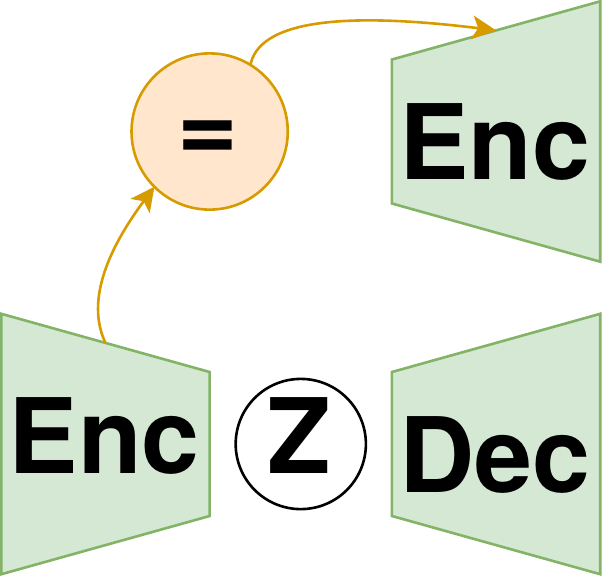}\end{tabular} & \refstepcounter{model}(\themodel)\label{model:m2}  encoder & 0.700 & 0.604\\
\multicolumn{1}{l}{} & \multicolumn{1}{l}{}                                 & \multicolumn{1}{l}{}                                 
\end{tabular}}
\caption{(a) FAVAE reconstruction and anomaly localization on MVTec AD.
Columns from left to right: [1] input image, [2] VAE reconstruction,
[3-6] \{pixel, VGG 1st, VGG 2nd, VGG 3rd\} layer log reconstruction probability,
[7] merged log probability of layers, [8] input image with
anomaly map overlay, [9] anomaly map with ground truth defect segmentation (white mask) overlay. (b) FAVAE Perceptual feature study on MVTec AD.}
\label{fig:mvtec_qualitative}
\end{figure}

\Figref{fig:mvtec_qualitative} presents qualitative results on MVTec AD. The anomaly map is obtained by equalizing the anomaly score histogram over the entire test set of each dataset. After equalization, a \textit{jet} colormap is applied.

Unless specified we used the same architecture for the encoder and decoder (detailed in Appendix 4) for all our experiments as well as the same hyperparameters when it applies. We did not do any hyperparameter optimization for any model.
Experiments were run 5 times per dataset and model.Each experiment takes about one hour (100 epochs) on a GTX 1080 Ti.

\subsection{Results}

In Appendix 5, we test several feature extractors.
Here we report the results for FAVAE augmented with VGG16 features.
We model the input of the 2nd, 3rd and 4th max pooling (right before activation)
Results are listed in \tabref{table:backbone}. The Vanilla VAE is our baseline model that targets only pixel reconstruction. We also added scores from \citet{Dehaene2020} who tested several anomaly localization algorithms. For each dataset we reported the \textit{best} score reported in their work. 
The corresponding model is either an autoencoder, a SSIM-autoencoder \citep{Bergmann2018}, a VAE or a $\gamma$-VAE \citep{Dai2019}, or either one of these augmented with AE-grad \citep{Dehaene2020}.

On the localization task (Pixel AUROC), FAVAE improves the \textit{Ref} scores from \citet{Dehaene2020}, which involve 8 different models and a comparatively slow gradient-based method, on all subset of MVTec AD.
On both detection and localization tasks the improvement over vanilla VAE and \textit{Ref} are significant.

\subsection{Study on perceptual features}


In the next experiment, we make several changes to the architecture to assess the benefits of the FAVAE approach. For example, does FAVAE work because we add structured, multi scale features or because those features really bias the model towards human perception ?
\Figref{fig:mvtec_qualitative}b presents the performance of 6 variants, averaged over all MVTec AD datasets.

(M\ref{model:m4}) is the standard pre-trained VGG16 FAVAE.
(M\ref{model:m5})'s feature extractor is a VGG16 backbone with randomly initialized weights.
(M\ref{model:m3}) uses its own encoder as a feature extractor with a gradient stop to prevent degenerate solutions.
(M\ref{model:m1}) is the vanilla VAE (no feature extractor).
(M\ref{model:m6}) is a VGG16 FAVAE like (M\ref{model:m4}) with unfrozen weights in the feature extractor.
(M\ref{model:m2}) is the same as (M\ref{model:m3}) without the gradient stop.

FAVAE (M\ref{model:m4}) with VGG16 as a feature extractor is the winning architecture both for localization and detection, which means that we both need high level features and perceptual features coming from ImageNet training for visual anomaly detection.
Interestingly (M\ref{model:m5}) with a randomly initialized VGG backbone and (M\ref{model:m3}) decoding its own encoder's features perform better than the vanilla VAE (M\ref{model:m1}) in localization but not in detection.
This may hint that we could better merge the anomaly maps of different layers.
The performance of (M\ref{model:m5}) and (M\ref{model:m3}) for localization also shows that adding multi scale features helps for anomaly localization. 
Finally, (M\ref{model:m2}) and (M\ref{model:m6}) offer poor performances as their feature extractors are allowed to generate useless and easy to predict features.

This second experiment comforts our hypothesis. 
It helps to model not only pixels but also higher-level
feature (even random) for the task of anomaly
localization ((M\ref{model:m4}), (M\ref{model:m5})
and (M\ref{model:m3}) performs better that vanilla
VAE).
Classifiers trained on ImageNet learn features
that transfer well ((M\ref{model:m4}) performs better than (M\ref{model:m5}) and (M\ref{model:m3})).

\section{Conclusion}

The difference between an outlier and an anomaly with respect to a normal distribution is that an outlier is supposed to come from the distribution but is a very rare event.
An anomaly, on the other hand, is a sample that is more probably coming from another distribution than the normal one.
In this paper we have shown on our toy dataset the interest of going beyond outlier detection for anomaly detection: in some cases we have to represent models of the anomalies to yield useful results.
Next, we have implicitly modeled richer and more perceptual anomalies by transferring features from pre-trained models and modeling these features' distributions using VAEs.
We have demonstrated that the features not only need to be of higher level than pixels, but also come from models pre-trained on large image datasets in order to gain maximum performance.
Finally, we showed that by simply augmenting a VAE with features from the VGG16 model trained on ImageNet, we outperformed state-of-the-art methods for anomaly detection and localization.

Future work could focus on refining the method for merging errors from different layers, and on the extension of this work to more complicated video datasets, using spatio-temporal perceptual features.
\clearpage
\bibliography{favae_paper}
\bibliographystyle{unsrtnat}
\clearpage
\appendix
\section{Formal definition of the toy dataset}
The normal distribution is defined as the following model:
$$q_n(\mu, \rvx) =  q_n(\mu) \: \prod_{k=1}^{d} q_n(\rx_k|\mu)$$
$$\forall k, \: q_n(\rx_{k}|\mu) \equiv \mathcal{N}(\mu, \sigma_{n}^{2}), \;\;\;\; q_n(\mu) \equiv \mathcal{N}(0, 1)$$
where $k$ indexes all pixels and $\sigma_n$ is a constant.

The anomaly distribution is defined as:
$$\forall i,j, \: \rx_{i,j} = \mu + G \sin \left(2\pi \psi j + \phi \right)$$
$$q_{a}(\phi, \mu, G) = q_a(\phi) \, q_{a}(\mu) \, q_{a}(G) $$
$$q_a(\mu) \equiv \mathcal{N}(0, 1), \;\; q_a(G) \equiv \mathcal{N}(0, \sigma_{a}^{2}), \;\; q_a(\phi) \equiv \mathcal{U}(0, 2\pi)$$
where $i$, $j$ respectively index pixels by columns and rows, $\psi$ and $\sigma_a$ are constants, $\sigma_a \text{ s.t. Var}[q_a(\rx_{i,j}|\mu, G, \phi)] < \sigma_n$.
This distribution has an underlying structure of horizontal stripes, with added white noise of an amplitude smaller than the noise amplitude of the normal distribution.

For the experiment we chose  $\sigma_n=0.0285$, $\sigma_a=0.0570$, $\psi=5$, $d=128 \times 128$.


\section{Proof that VAE is optimal on the toy dataset}

We want to prove that the VAE will learn $\theta$ so that $p_{\theta}(\rvx) = q_n(\rvx)$ by optimising the ELBO.
A known \citep{kingma2013auto} form of the ELBO is:
\begin{equation}
    \text{ELBO}_{\theta, \phi} = - \mathcal{H}(q_n(\rvx)) - \KL(q_n(\rvx)\Vert p_{\theta}(\rvx)) - \E_{q_n(\rvx)}\KL(q_{\phi}(\rvz|\rvx)\Vert p_{\theta}(\rvz|\rvx))
\label{eq:ELBO_form}
\end{equation}
with $\mathcal{H}$ the entropy. The KL divergence is positive and the entropy term is constant with respect to $\theta, \phi$, so that iff $\KL(q_n(\rvx)\Vert p_{\theta}(\rvx)) = 0$ and $\E_{q_n(\rvx)} \KL(q_{\phi}(\rvz|\rvx)\Vert p_{\theta}(\rvz|\rvx)) = 0$, the ELBO is maximized with respect to $\theta, \phi$.

If $$p_{\theta}(\rvx|\rvz) = \prod_{k=1}^d p_{\theta}(\rx_k|\rvz)$$ 
with 
$$\forall k, p_{\theta}(\rx_k|\rvz) \equiv \mathcal{N} ( Id(\rvz), \sigma_n^2 \,)$$ and with the standard VAE prior $p(\rvz) = \mathcal{N}(0, 1)$, then we have $p_{\theta}(\rvx) = q_n(\rvx)$.

The decoder simply has to broadcast the latent variable $\rvz$ to all pixels: $$\forall k, \mu_{\theta, k}(\rvz) = \rvz$$ and the learnable parameter of the decoder variance must reach:
$$\gamma = \sigma_n$$
so that $\KL(q_n(\rvx)\Vert p_{\theta}(\rvx)) = 0$  in \eqref{eq:ELBO_form}. We suppose the decoder architecture capable of implementing this very simple solution, and we call $\theta^*$ its corresponding parameters.

We can now prove that there exists $\phi$ so that $\KL(q_{\phi}(\rvz|\rvx)\Vert p_{\theta^*}(\rvz|\rvx)) = 0$.
The Gaussian family is self-conjugate with respect to a Gaussian likelihood function \citep{wiki:conjugate}, which means, considering the pixels {$\rx_1,.. \rx_d$} as a set of observations of the likelihood function $p_{\theta^*}(\rx_k|\rvz)$, that the posterior $p_{\theta^*}(\rvz|\rvx)$ is Gaussian and that:
\begin{equation}
    p_{\theta^*}(\rvz|\rvx) = \mathcal{N} \left( \frac{\sum_{k=1}^d \rx_k}{\sigma_n^2 + d} \, , \, \frac{\sigma_n^2}{\sigma_n^2 + d} \right)
\end{equation}
We suppose the encoder architecture able to implement this simple solution, and we call $\phi^*$ its corresponding parameters.

We have thus found $\theta^*, \phi^*$ maximizing the ELBO and so that $\KL(q_n(\rvx)\Vert p_{\theta^*}(\rvx)) = 0$. By optimizing the ELBO, the VAE can find the optimal generative model $p_{\theta^*}(\rvx) = q_n(\rvx)$.
\section{Implicit anomaly models in standard approaches}
\subsection{Models based on encoder-decoder reconstruction}
We will show how to get the standard approach pixel-wise normal/anomalous classifier 
\begin{equation}
\forall k, \;\; (C_{a_k}(\rvx) > 0.5) \iff (p_{\theta}(\rx_k|\vz_{\phi}(\rvx)) < T)
\label{eq:classic_classifier}
\end{equation}
from the following anomaly sampling process:
\begin{equation}
\begin{split}
\rvx_a &= \mu_{\theta}(\rvz) + \ve \\
\rvz \sim p(\rvz), \;\;\;\; &\ve \sim p_a(\ve) \equiv \mathcal{N}(0, 
\sigma_a^2)
\end{split}
\label{eq:classic_anomaly_model}
\end{equation}
with $\sigma_a$ a constant.

We will first need to explicit approximations that are commonly used with the VAE. We suppose that since from \eqref{eq:ELBO_form} $\KL(q_{\phi}(\rvz|\rvx) \Vert p_{\theta}(\rvz|\rvx))$ is minimised on the training set, we can approximate $p_{\theta}(\rvz|\rvx)$ by $q_{\phi}(\rvz|\rvx)$ for $\rvx \sim q_n(\rvx)$. We also suppose that the encoded distributions $q_{\phi}(\rvz|\rvx)$ are sufficiently narrow, so that we can only decode the mean of the encoded distribution, $\vz_{\phi}(\rvx)$.
These two approximations correspond to only taking into account the quality of the VAE reconstruction.
Another common approximation that is typical to the standard VAE-based anomaly detection framework is to suppose that $p_{\theta}(\rvz|\rvx) \approx q_{\phi}(\rvz|\rvx)$ remains true  even for $\rvx$ sampled from another distribution than the normal distribution.
This corresponds to using an encoder trained on the normal distribution on the anomalous distribution even though we know they are different.
These approximations enable us to write:
\begin{equation}
p_{\theta}(a|\rvx) = \int p_{\theta}(a|\rvx, \rvz) \, p_{\theta}(\rvz|\rvx) \, d\rvz \approx \int p_{\theta}(a|\rvx, \rvz) \, q_{\phi}(\rvz|\rvx) \, d\rvz \approx p_{\theta}(a|\rvx, \vz_{\phi}(\rvx))
\label{eq:classic_approximations}
\end{equation}
\begin{equation}
p_{\theta}(n|\rvx) = \int p_{\theta}(n|\rvx, \rvz) \, p_{\theta}(\rvz|\rvx) \, d\rvz \approx \int p_{\theta}(n|\rvx, \rvz) \, q_{\phi}(\rvz|\rvx) \, d\rvz \approx p_{\theta}(n|\rvx, \vz_{\phi}(\rvx))
\label{eq:classic_approximations}
\end{equation}
Then
\begin{equation}
C_{a_k}(\rvx) = \frac{p_{\theta}(a|\rvx)}{p_{\theta}(n|\rvx)} 
\approx \frac{p_{\theta}(a|\rx_k, \vz_{\phi}(\rvx))}{p_{\theta}(n|\rx_k, \vz_{\phi}(\rvx))}
\approx \frac{p_{a, \theta}(\rx_k|\vz_{\phi}(\rvx))p(a)}{p_{n, \theta}(\rx_k|\vz_{\phi}(\rvx))p(n)}
\end{equation}
Since $p_a(\rx_k|\rvz)$ is a Gaussian of same mean but different scale as $p_{\theta}(\rx_k|\rvz)$, they intersect at two points that are symmetric with respect to their mean value, and so we get \eqref{eq:classic_classifier} as a result.

\subsection{The recent model from \citet{Dehaene2020}}
We showed that classical VAE-based anomaly detection models implicitly used simplistic, pixel-wise independent anomaly models that could hinder their practicality.
In this section we will show as another example that the more recent approach of \citet{Dehaene2020}, while it does not need some aforementioned approximations, also uses the same kind of pixel-wise anomaly models.

They propose to search for an optimal "corrected" image $\vx_t$ on the normal data manifold using gradient descent on the objective
$$\mathcal{L}_r(\vx_t) + \lambda||\vx_t - \rvx||_1$$
with $\mathcal{L}_r(\vx_t)$ the VAE reconstruction loss.
They derive this objective from an adversarial examples framework, however we show here that it can also emerge from an explicit model of the anomaly distribution. 
Let us assume that the test image is the composition of a sample from the normal distribution and white pixel noise from a Laplace distribution 
$$\rvx = \vx_n + \ve$$
$$\vx_n \sim q_n(\rx_n) \;\;\; \ve \sim q_e(\ve) \equiv Lap(0, 1/\lambda)$$
we search for the most probable pair of realizations $(\rvx, \vx_n)$ according to that model, i.e. the $\vx_n$ that minimizes
\begin{equation}
\begin{split}
-\log q(\rvx, \vx_n) &=  - \log ( \, q(\vx_n) \, q(\rvx|\vx_n) \, ) \\
&= -\log q_n(\vx_n) - \log q_e(\ve) \\
&= -\log q_n(\vx_n) - \log q_e(\rvx - \vx_n) \\
&\approx - ELBO(\vx_n) + \lambda||\vx_n - \rvx||_1 + K \\
&\approx \mathcal{L}_r(\vx_n) + \lambda||\vx_n - \rvx||_1 + K
\end{split}
\end{equation}
K is as constant with respect to $\vx_n$.
We can thus see that even though they add a more flexible model, anomalies are still considered as independent noise over pixels.
\section{Architecture}

This section presents the architecture of the modules used to implement FAVAE. We reproduce here the Pytorch string representation of the main modules.

The encoder is a Fully Convolutional Network. 
The output of the encoder is a 200x1x1 tensor for a 3x128x128 input image.
The output tensor is split into two parts (mean and sigma of the encoded distribution). 
Thus the latent space dimension is 100.

\begin{multicols}{2}

\lstset{
    basicstyle=\footnotesize\ttfamily, 
    numberstyle=\tiny,          
    numbersep=5pt,              
    tabsize=2,                  
    extendedchars=true,
    breaklines=true,            
    keywordstyle=\color{red},
    stringstyle=\color{white}\ttfamily, 
    showspaces=false,
    showtabs=false,
    xleftmargin=17pt,
    framexleftmargin=17pt,
    framexrightmargin=5pt,
    framexbottommargin=4pt,
    showstringspaces=false
}

\begin{lstlisting}
Sequential(
  (0): Conv2d(3, 128,
  		  kernel_size=(4, 4),
          stride=(2, 2),
          padding=(1, 1))
  (1): BatchNorm2d(128,
          eps=1e-05,
          momentum=0.1,
          affine=True,
          track_running_stats=True)
  (2): LeakyReLU(negative_slope=0.2)
  (3): Conv2d(128, 128,
          kernel_size=(4, 4),
          stride=(2, 2),
          padding=(1, 1))
  (4): BatchNorm2d(128,
          eps=1e-05,
          momentum=0.1,
          affine=True,
          track_running_stats=True)
  (5): LeakyReLU(negative_slope=0.2)
  (6): Conv2d(128, 256,
          kernel_size=(3, 3),
          stride=(1, 1),
          padding=(1, 1))
  (7): BatchNorm2d(256,
          eps=1e-05,
          momentum=0.1,
          affine=True,
          track_running_stats=True)
  (8): LeakyReLU(negative_slope=0.2)
  (9): Conv2d(256, 256,
          kernel_size=(4, 4),
          stride=(2, 2),
          padding=(1, 1))
  (10): BatchNorm2d(256,
          eps=1e-05,
          momentum=0.1,
          affine=True,
          track_running_stats=True)
  (11): LeakyReLU(negative_slope=0.2)
  (12): Conv2d(256, 512,
          kernel_size=(3, 3),
          stride=(1, 1),
          padding=(1, 1))
  (13): BatchNorm2d(512,
          eps=1e-05,
          momentum=0.1,
          affine=True,
          track_running_stats=True)
  (14): LeakyReLU(negative_slope=0.2)
  (15): Conv2d(512, 512,
          kernel_size=(4, 4),
          stride=(2, 2),
          padding=(1, 1))
  (16): BatchNorm2d(512,
          eps=1e-05,
          momentum=0.1,
          affine=True,
          track_running_stats=True)
  (17): LeakyReLU(negative_slope=0.2)
  (18): Conv2d(512, 512,
          kernel_size=(3, 3),
          stride=(1, 1),
          padding=(1, 1))
  (19): BatchNorm2d(512,
          eps=1e-05,
          momentum=0.1, 
          affine=True,
          track_running_stats=True)
  (20): LeakyReLU(negative_slope=0.2)
  (21): Conv2d(512, 32,
          kernel_size=(3, 3),
          stride=(1, 1),
          padding=(1, 1))
  (22): BatchNorm2d(32,
          eps=1e-05,
          momentum=0.1,
          affine=True,
          track_running_stats=True)
  (23): LeakyReLU(negative_slope=0.2)
  (24): Conv2d(32, 200,
          kernel_size=(8, 8),
          stride=(1, 1))
  (25): Flatten()
  (26): Split()
)
\end{lstlisting}
\end{multicols}

The decoder is the symmetric of the encoder, using transposed convolutions.
\begin{multicols}{2}
\begin{lstlisting}
Sequential(
  (0): DeFlatten(shape=(100, 1, 1))
  (1): ConvTranspose2d(100, 32,
          kernel_size=(8, 8),
          stride=(1, 1),
          output_padding=[0, 0])
  (2): BatchNorm2d(32,
          eps=1e-05,
          momentum=0.1,
          affine=True,
          track_running_stats=True)
  (3): LeakyReLU(negative_slope=0.2)
  (4): ConvTranspose2d(32, 512,
          kernel_size=(3, 3),
          stride=(1, 1),
          padding=(1, 1),
          output_padding=[0, 0])
  (5): BatchNorm2d(512,
          eps=1e-05,
          momentum=0.1,
          affine=True,
          track_running_stats=True)
  (6): LeakyReLU(negative_slope=0.2)
  (7): ConvTranspose2d(512, 512, 
          kernel_size=(3, 3),
          stride=(1, 1),
          padding=(1, 1),
          output_padding=[0, 0])
  (8): BatchNorm2d(512,
          eps=1e-05,
          momentum=0.1,
          affine=True,
          track_running_stats=True)
  (9): LeakyReLU(negative_slope=0.2)
  (10): ConvTranspose2d(512, 512,
          kernel_size=(4, 4),
          stride=(2, 2),
          padding=(1, 1),
          output_padding=[0, 0])
  (11): BatchNorm2d(512,
          eps=1e-05,
          momentum=0.1,
          affine=True,
          track_running_stats=True)
  (12): LeakyReLU(negative_slope=0.2)
  (13): ConvTranspose2d(512, 256,
          kernel_size=(3, 3),
          stride=(1, 1),
          padding=(1, 1),
          output_padding=[0, 0])
  (14): BatchNorm2d(256,
          eps=1e-05,
          momentum=0.1,
          affine=True,
          track_running_stats=True)
  (15): LeakyReLU(negative_slope=0.2)
  (16): ConvTranspose2d(256, 256,
          kernel_size=(4, 4),
          stride=(2, 2),
          padding=(1, 1),
          output_padding=[0, 0])
  (17): BatchNorm2d(256,
          eps=1e-05,
          momentum=0.1,
          affine=True,
          track_running_stats=True)
  (18): LeakyReLU(negative_slope=0.2)
  (19): ConvTranspose2d(256, 128,
          kernel_size=(3, 3),
          stride=(1, 1),
          padding=(1, 1),
          output_padding=[0, 0])
  (20): BatchNorm2d(128,
          eps=1e-05,
          momentum=0.1,
          affine=True,
          track_running_stats=True)
  (21): LeakyReLU(negative_slope=0.2)
  (22): ConvTranspose2d(128, 128,
          kernel_size=(4, 4),
          stride=(2, 2),
          padding=(1, 1),
          output_padding=[0, 0])
  (23): BatchNorm2d(128,
          eps=1e-05,
          momentum=0.1,
          affine=True,
          track_running_stats=True)
  (24): LeakyReLU(negative_slope=0.2)
  (25): ConvTranspose2d(128, 3,
          kernel_size=(4, 4),
          stride=(2, 2),
          padding=(1, 1),
          output_padding=[0, 0])
  (26): Identity()
  (27): Sigmoid()
        )
\end{lstlisting}
\end{multicols}{}

Below is an example of the 3 feature adapters for the VGG case. 
These adapters are 2-layers CNNs with a 1x1 kernel.
The hidden layer and the output layer sizes are set to the number of channels of the
feature extractor layer to match.
\begin{multicols}{2}
\begin{lstlisting}
ModuleList(
  (0): Sequential(
    (0): Conv2d(128, 128,
    kernel_size=(1, 1),
     stride=(1, 1))
    (1): ReLU()
    (2): Conv2d(128, 128,
    kernel_size=(1, 1),
     stride=(1, 1))
  )
  (1): Sequential(
    (0): Conv2d(256, 256,
    kernel_size=(1, 1),
     stride=(1, 1))
    (1): ReLU()
    (2): Conv2d(256, 256,
    kernel_size=(1, 1),
     stride=(1, 1))
  )
\end{lstlisting}  
\begin{lstlisting}
  (2): Sequential(
    (0): Conv2d(512, 512, 
    kernel_size=(1, 1),
     stride=(1, 1))
    (1): ReLU()
    (2): Conv2d(512, 512, 
    kernel_size=(1, 1),
     stride=(1, 1))
  )
)
\end{lstlisting}

\end{multicols}

Finally \tabref{table:connections} summarizes the connections between the decoder and the feature extractor.
\begin{table}[h!]
\begin{center}
\begin{tabular}{|c|c|c|}
    \hline
  	\multicolumn{1}{|c|}{\begin{tabular}[c]{@{}c@{}}\textbf{Feature}\\\textbf{Extractor}\end{tabular}} & \multicolumn{1}{c|}{\begin{tabular}[c]{@{}c@{}}\textbf{Feature}\\\textbf{Layers}\end{tabular}} & \multicolumn{1}{c|}{\begin{tabular}[c]{@{}c@{}}\textbf{Decoder}\\\textbf{Layers}\end{tabular}} \\ \hline
	VGG16 & (7, 14, 21) & (22, 10, 16) \\ \hline
	Resnet18 & (conv2x, conv3x, conv4x) & (22, 10, 16) \\ \hline
	YOLOv3 & 20 & 10 \\ \hline
	\begin{tabular}[c]{@{}c@{}}auto feature\\extractor\end{tabular} & (3,9,15) & (22, 10, 16) \\ \hline
	
\end{tabular}
\end{center}
\caption{Reconstruction mapping between decoder layers and feature extractor layers.}
\label{table:connections}
\end{table}

\section{Which feature extractor ?}

\definecolor{gray}{rgb}{0.9, 0.9, 0.9}

\begin{table*}
\caption{MVTec result for several feature extractors. \textit{Ref} are the best results from \citet{Dehaene2020}. A green cell indicates that the VGG16 FAVAE model performs better than vanilla VAE (on average) for a given dataset and metric. Best result for a given dataset and metric is written in bold. Each experiment has been run 5 times. We report mean and standard deviation for each experiment.}

\newcommand{\doublecell}[1]{\multirow{2}{1.3cm}{\centering \textbf{#1}}}
\begin{center}
\resizebox{\columnwidth}{!}{
 \begin{tabular}{|c|c|c|c|c|c||c|c|c|c|}
    \hline
  	\multirow{3}{1cm}{\textbf{Dataset}} &
    \multicolumn{5}{c||}{\textbf{Pixel AUROC}} &
    \multicolumn{4}{c|}{\textbf{Image AUROC}} \\ \cline{2-10}
    & \doublecell{Ref} & \doublecell{VGG16 FAVAE} & \doublecell{Vanilla VAE} & \doublecell{YOLOv3 FAVAE} & \doublecell{Resnet18 FAVAE} 
    & \doublecell{Vanilla VAE} & \doublecell{VGG16 FAVAE} & \doublecell{YOLOv3 FAVAE} & \doublecell{Resnet18 FAVAE}  \\ &&&&&&&&& \\ \hline
 carpet &  \begin{tabular}[c]{@{}c@{}} \vspace{-2mm} 0.774\\ {\tiny - } \end{tabular} & \green \begin{tabular}[c]{@{}c@{}}\textbf{ \vspace{-2mm} 0.960}\\ {\tiny $\pm$ 0.004}\end{tabular} &  \begin{tabular}[c]{@{}c@{}} \vspace{-2mm} 0.620\\ {\tiny $\pm$ 0.006}\end{tabular} &  \begin{tabular}[c]{@{}c@{}} \vspace{-2mm} 0.847\\ {\tiny $\pm$ 0.014}\end{tabular} &  \begin{tabular}[c]{@{}c@{}} \vspace{-2mm} 0.910\\ {\tiny $\pm$ 0.005}\end{tabular} &  \begin{tabular}[c]{@{}c@{}} \vspace{-2mm} 0.515\\ {\tiny $\pm$ 0.011}\end{tabular} & \green \begin{tabular}[c]{@{}c@{}} \vspace{-2mm} 0.671\\ {\tiny $\pm$ 0.010}\end{tabular} &  \begin{tabular}[c]{@{}c@{}} \vspace{-2mm} 0.629\\ {\tiny $\pm$ 0.006}\end{tabular} &  \begin{tabular}[c]{@{}c@{}}\textbf{ \vspace{-2mm} 0.806}\\ {\tiny $\pm$ 0.008}\end{tabular}\\ \hline 
 grid &  \begin{tabular}[c]{@{}c@{}} \vspace{-2mm} 0.981\\ {\tiny - } \end{tabular} & \green \begin{tabular}[c]{@{}c@{}}\textbf{ \vspace{-2mm} 0.993}\\ {\tiny $\pm$ 0.000}\end{tabular} &  \begin{tabular}[c]{@{}c@{}} \vspace{-2mm} 0.856\\ {\tiny $\pm$ 0.003}\end{tabular} &  \begin{tabular}[c]{@{}c@{}} \vspace{-2mm} 0.988\\ {\tiny $\pm$ 0.000}\end{tabular} &  \begin{tabular}[c]{@{}c@{}} \vspace{-2mm} 0.941\\ {\tiny $\pm$ 0.004}\end{tabular} &  \begin{tabular}[c]{@{}c@{}} \vspace{-2mm} 0.904\\ {\tiny $\pm$ 0.037}\end{tabular} & \green \begin{tabular}[c]{@{}c@{}}\textbf{ \vspace{-2mm} 0.970}\\ {\tiny $\pm$ 0.004}\end{tabular} &  \begin{tabular}[c]{@{}c@{}} \vspace{-2mm} 0.854\\ {\tiny $\pm$ 0.032}\end{tabular} &  \begin{tabular}[c]{@{}c@{}} \vspace{-2mm} 0.767\\ {\tiny $\pm$ 0.019}\end{tabular}\\ \hline 
 leather &  \begin{tabular}[c]{@{}c@{}} \vspace{-2mm} 0.925\\ {\tiny - } \end{tabular} & \green \begin{tabular}[c]{@{}c@{}} \vspace{-2mm} 0.981\\ {\tiny $\pm$ 0.001}\end{tabular} &  \begin{tabular}[c]{@{}c@{}} \vspace{-2mm} 0.835\\ {\tiny $\pm$ 0.015}\end{tabular} &  \begin{tabular}[c]{@{}c@{}} \vspace{-2mm} 0.969\\ {\tiny $\pm$ 0.011}\end{tabular} &  \begin{tabular}[c]{@{}c@{}}\textbf{ \vspace{-2mm} 0.982}\\ {\tiny $\pm$ 0.001}\end{tabular} &  \begin{tabular}[c]{@{}c@{}} \vspace{-2mm} 0.619\\ {\tiny $\pm$ 0.043}\end{tabular} & \green \begin{tabular}[c]{@{}c@{}} \vspace{-2mm} 0.675\\ {\tiny $\pm$ 0.013}\end{tabular} &  \begin{tabular}[c]{@{}c@{}} \vspace{-2mm} 0.878\\ {\tiny $\pm$ 0.017}\end{tabular} &  \begin{tabular}[c]{@{}c@{}}\textbf{ \vspace{-2mm} 0.898}\\ {\tiny $\pm$ 0.005}\end{tabular}\\ \hline 
 tile &  \begin{tabular}[c]{@{}c@{}} \vspace{-2mm} 0.654\\ {\tiny - } \end{tabular} & \green \begin{tabular}[c]{@{}c@{}} \vspace{-2mm} 0.714\\ {\tiny $\pm$ 0.001}\end{tabular} &  \begin{tabular}[c]{@{}c@{}} \vspace{-2mm} 0.520\\ {\tiny $\pm$ 0.005}\end{tabular} &  \begin{tabular}[c]{@{}c@{}}\textbf{ \vspace{-2mm} 0.867}\\ {\tiny $\pm$ 0.006}\end{tabular} &  \begin{tabular}[c]{@{}c@{}} \vspace{-2mm} 0.857\\ {\tiny $\pm$ 0.003}\end{tabular} &  \begin{tabular}[c]{@{}c@{}} \vspace{-2mm} 0.512\\ {\tiny $\pm$ 0.030}\end{tabular} & \green \begin{tabular}[c]{@{}c@{}} \vspace{-2mm} 0.805\\ {\tiny $\pm$ 0.008}\end{tabular} &  \begin{tabular}[c]{@{}c@{}}\textbf{ \vspace{-2mm} 0.850}\\ {\tiny $\pm$ 0.011}\end{tabular} &  \begin{tabular}[c]{@{}c@{}} \vspace{-2mm} 0.832\\ {\tiny $\pm$ 0.003}\end{tabular}\\ \hline  
 wood &  \begin{tabular}[c]{@{}c@{}} \vspace{-2mm} 0.838\\ {\tiny - } \end{tabular} & \green \begin{tabular}[c]{@{}c@{}}\textbf{ \vspace{-2mm} 0.899}\\ {\tiny $\pm$ 0.002}\end{tabular} &  \begin{tabular}[c]{@{}c@{}} \vspace{-2mm} 0.699\\ {\tiny $\pm$ 0.002}\end{tabular} &  \begin{tabular}[c]{@{}c@{}} \vspace{-2mm} 0.879\\ {\tiny $\pm$ 0.008}\end{tabular} &  \begin{tabular}[c]{@{}c@{}} \vspace{-2mm} 0.833\\ {\tiny $\pm$ 0.009}\end{tabular} &  \begin{tabular}[c]{@{}c@{}} \vspace{-2mm} 0.875\\ {\tiny $\pm$ 0.009}\end{tabular} & \green \begin{tabular}[c]{@{}c@{}}\textbf{ \vspace{-2mm} 0.948}\\ {\tiny $\pm$ 0.002}\end{tabular} &  \begin{tabular}[c]{@{}c@{}} \vspace{-2mm} 0.882\\ {\tiny $\pm$ 0.032}\end{tabular} &  \begin{tabular}[c]{@{}c@{}} \vspace{-2mm} 0.877\\ {\tiny $\pm$ 0.004}\end{tabular}\\ \hline 
 bottle &  \begin{tabular}[c]{@{}c@{}} \vspace{-2mm} 0.951\\ {\tiny - } \end{tabular} & \green \begin{tabular}[c]{@{}c@{}}\textbf{ \vspace{-2mm} 0.963}\\ {\tiny $\pm$ 0.000}\end{tabular} &  \begin{tabular}[c]{@{}c@{}} \vspace{-2mm} 0.894\\ {\tiny $\pm$ 0.002}\end{tabular} &  \begin{tabular}[c]{@{}c@{}} \vspace{-2mm} 0.939\\ {\tiny $\pm$ 0.001}\end{tabular} &  \begin{tabular}[c]{@{}c@{}} \vspace{-2mm} 0.939\\ {\tiny $\pm$ 0.001}\end{tabular} &  \begin{tabular}[c]{@{}c@{}} \vspace{-2mm} 0.985\\ {\tiny $\pm$ 0.002}\end{tabular} & \green \begin{tabular}[c]{@{}c@{}}\textbf{ \vspace{-2mm} 0.999}\\ {\tiny $\pm$ 0.001}\end{tabular} &  \begin{tabular}[c]{@{}c@{}} \vspace{-2mm} 0.990\\ {\tiny $\pm$ 0.001}\end{tabular} &  \begin{tabular}[c]{@{}c@{}} \vspace{-2mm} 0.952\\ {\tiny $\pm$ 0.001}\end{tabular}\\ \hline 
 cable &  \begin{tabular}[c]{@{}c@{}} \vspace{-2mm} 0.910\\ {\tiny - } \end{tabular} & \green \begin{tabular}[c]{@{}c@{}}\textbf{ \vspace{-2mm} 0.969}\\ {\tiny $\pm$ 0.001}\end{tabular} &  \begin{tabular}[c]{@{}c@{}} \vspace{-2mm} 0.816\\ {\tiny $\pm$ 0.014}\end{tabular} &  \begin{tabular}[c]{@{}c@{}} \vspace{-2mm} 0.937\\ {\tiny $\pm$ 0.001}\end{tabular} &  \begin{tabular}[c]{@{}c@{}} \vspace{-2mm} 0.916\\ {\tiny $\pm$ 0.002}\end{tabular} &  \begin{tabular}[c]{@{}c@{}} \vspace{-2mm} 0.805\\ {\tiny $\pm$ 0.008}\end{tabular} & \green \begin{tabular}[c]{@{}c@{}}\textbf{ \vspace{-2mm} 0.950}\\ {\tiny $\pm$ 0.003}\end{tabular} &  \begin{tabular}[c]{@{}c@{}} \vspace{-2mm} 0.928\\ {\tiny $\pm$ 0.007}\end{tabular} &  \begin{tabular}[c]{@{}c@{}} \vspace{-2mm} 0.811\\ {\tiny $\pm$ 0.009}\end{tabular}\\ \hline 
 capsule &  \begin{tabular}[c]{@{}c@{}} \vspace{-2mm} 0.952\\ {\tiny - } \end{tabular} & \green \begin{tabular}[c]{@{}c@{}}\textbf{ \vspace{-2mm} 0.976}\\ {\tiny $\pm$ 0.001}\end{tabular} &  \begin{tabular}[c]{@{}c@{}} \vspace{-2mm} 0.907\\ {\tiny $\pm$ 0.010}\end{tabular} &  \begin{tabular}[c]{@{}c@{}} \vspace{-2mm} 0.949\\ {\tiny $\pm$ 0.002}\end{tabular} &  \begin{tabular}[c]{@{}c@{}} \vspace{-2mm} 0.947\\ {\tiny $\pm$ 0.003}\end{tabular} &  \begin{tabular}[c]{@{}c@{}} \vspace{-2mm} 0.694\\ {\tiny $\pm$ 0.038}\end{tabular} & \green \begin{tabular}[c]{@{}c@{}}\textbf{ \vspace{-2mm} 0.804}\\ {\tiny $\pm$ 0.008}\end{tabular} &  \begin{tabular}[c]{@{}c@{}} \vspace{-2mm} 0.717\\ {\tiny $\pm$ 0.015}\end{tabular} &  \begin{tabular}[c]{@{}c@{}} \vspace{-2mm} 0.682\\ {\tiny $\pm$ 0.011}\end{tabular}\\ \hline 
 hazelnut &  \begin{tabular}[c]{@{}c@{}}\textbf{ \vspace{-2mm} 0.988}\\ {\tiny - } \end{tabular} & \green \begin{tabular}[c]{@{}c@{}} \vspace{-2mm} 0.987\\ {\tiny $\pm$ 0.000}\end{tabular} &  \begin{tabular}[c]{@{}c@{}} \vspace{-2mm} 0.951\\ {\tiny $\pm$ 0.002}\end{tabular} &  \begin{tabular}[c]{@{}c@{}} \vspace{-2mm} 0.967\\ {\tiny $\pm$ 0.001}\end{tabular} &  \begin{tabular}[c]{@{}c@{}} \vspace{-2mm} 0.970\\ {\tiny $\pm$ 0.002}\end{tabular} &  \begin{tabular}[c]{@{}c@{}} \vspace{-2mm} 0.922\\ {\tiny $\pm$ 0.016}\end{tabular} & \green \begin{tabular}[c]{@{}c@{}}\textbf{ \vspace{-2mm} 0.993}\\ {\tiny $\pm$ 0.001}\end{tabular} &  \begin{tabular}[c]{@{}c@{}} \vspace{-2mm} 0.982\\ {\tiny $\pm$ 0.003}\end{tabular} &  \begin{tabular}[c]{@{}c@{}} \vspace{-2mm} 0.940\\ {\tiny $\pm$ 0.006}\end{tabular}\\ \hline 
 metal nut &  \begin{tabular}[c]{@{}c@{}} \vspace{-2mm} 0.920\\ {\tiny - } \end{tabular} & \green \begin{tabular}[c]{@{}c@{}}\textbf{ \vspace{-2mm} 0.966}\\ {\tiny $\pm$ 0.001}\end{tabular} &  \begin{tabular}[c]{@{}c@{}} \vspace{-2mm} 0.861\\ {\tiny $\pm$ 0.009}\end{tabular} &  \begin{tabular}[c]{@{}c@{}} \vspace{-2mm} 0.951\\ {\tiny $\pm$ 0.001}\end{tabular} &  \begin{tabular}[c]{@{}c@{}} \vspace{-2mm} 0.924\\ {\tiny $\pm$ 0.003}\end{tabular} &  \begin{tabular}[c]{@{}c@{}} \vspace{-2mm} 0.676\\ {\tiny $\pm$ 0.010}\end{tabular} & \green \begin{tabular}[c]{@{}c@{}} \vspace{-2mm} 0.852\\ {\tiny $\pm$ 0.004}\end{tabular} &  \begin{tabular}[c]{@{}c@{}}\textbf{ \vspace{-2mm} 0.854}\\ {\tiny $\pm$ 0.015}\end{tabular} &  \begin{tabular}[c]{@{}c@{}} \vspace{-2mm} 0.788\\ {\tiny $\pm$ 0.009}\end{tabular}\\ \hline 
 pill &  \begin{tabular}[c]{@{}c@{}} \vspace{-2mm} 0.935\\ {\tiny - } \end{tabular} & \green \begin{tabular}[c]{@{}c@{}}\textbf{ \vspace{-2mm} 0.953}\\ {\tiny $\pm$ 0.001}\end{tabular} &  \begin{tabular}[c]{@{}c@{}} \vspace{-2mm} 0.879\\ {\tiny $\pm$ 0.003}\end{tabular} &  \begin{tabular}[c]{@{}c@{}} \vspace{-2mm} 0.948\\ {\tiny $\pm$ 0.001}\end{tabular} &  \begin{tabular}[c]{@{}c@{}} \vspace{-2mm} 0.945\\ {\tiny $\pm$ 0.001}\end{tabular} &  \begin{tabular}[c]{@{}c@{}} \vspace{-2mm} 0.808\\ {\tiny $\pm$ 0.008}\end{tabular} & \green \begin{tabular}[c]{@{}c@{}}\textbf{ \vspace{-2mm} 0.821}\\ {\tiny $\pm$ 0.009}\end{tabular} &  \begin{tabular}[c]{@{}c@{}} \vspace{-2mm} 0.773\\ {\tiny $\pm$ 0.013}\end{tabular} &  \begin{tabular}[c]{@{}c@{}} \vspace{-2mm} 0.750\\ {\tiny $\pm$ 0.010}\end{tabular}\\ \hline 
 screw &  \begin{tabular}[c]{@{}c@{}} \vspace{-2mm} 0.983\\ {\tiny - } \end{tabular} & \green \begin{tabular}[c]{@{}c@{}}\textbf{ \vspace{-2mm} 0.993}\\ {\tiny $\pm$ 0.003}\end{tabular} &  \begin{tabular}[c]{@{}c@{}} \vspace{-2mm} 0.928\\ {\tiny $\pm$ 0.028}\end{tabular} &  \begin{tabular}[c]{@{}c@{}} \vspace{-2mm} 0.938\\ {\tiny $\pm$ 0.004}\end{tabular} &  \begin{tabular}[c]{@{}c@{}} \vspace{-2mm} 0.937\\ {\tiny $\pm$ 0.002}\end{tabular} &  \begin{tabular}[c]{@{}c@{}} \vspace{-2mm} 0.654\\ {\tiny $\pm$ 0.111}\end{tabular} & \green \begin{tabular}[c]{@{}c@{}}\textbf{ \vspace{-2mm} 0.837}\\ {\tiny $\pm$ 0.101}\end{tabular} &  \begin{tabular}[c]{@{}c@{}} \vspace{-2mm} 0.452\\ {\tiny $\pm$ 0.043}\end{tabular} &  \begin{tabular}[c]{@{}c@{}} \vspace{-2mm} 0.476\\ {\tiny $\pm$ 0.035}\end{tabular}\\ \hline 
 toothbrush &  \begin{tabular}[c]{@{}c@{}} \vspace{-2mm} 0.985\\ {\tiny - } \end{tabular} & \green \begin{tabular}[c]{@{}c@{}}\textbf{ \vspace{-2mm} 0.987}\\ {\tiny $\pm$ 0.000}\end{tabular} &  \begin{tabular}[c]{@{}c@{}} \vspace{-2mm} 0.953\\ {\tiny $\pm$ 0.002}\end{tabular} &  \begin{tabular}[c]{@{}c@{}} \vspace{-2mm} 0.976\\ {\tiny $\pm$ 0.000}\end{tabular} &  \begin{tabular}[c]{@{}c@{}} \vspace{-2mm} 0.948\\ {\tiny $\pm$ 0.002}\end{tabular} &  \begin{tabular}[c]{@{}c@{}}\textbf{ \vspace{-2mm} 0.987}\\ {\tiny $\pm$ 0.004}\end{tabular} & \red \begin{tabular}[c]{@{}c@{}} \vspace{-2mm} 0.958\\ {\tiny $\pm$ 0.006}\end{tabular} &  \begin{tabular}[c]{@{}c@{}} \vspace{-2mm} 0.964\\ {\tiny $\pm$ 0.007}\end{tabular} &  \begin{tabular}[c]{@{}c@{}} \vspace{-2mm} 0.891\\ {\tiny $\pm$ 0.013}\end{tabular}\\ \hline 
 transistor &  \begin{tabular}[c]{@{}c@{}} \vspace{-2mm} 0.934\\ {\tiny - } \end{tabular} & \green \begin{tabular}[c]{@{}c@{}}\textbf{ \vspace{-2mm} 0.984}\\ {\tiny $\pm$ 0.001}\end{tabular} &  \begin{tabular}[c]{@{}c@{}} \vspace{-2mm} 0.851\\ {\tiny $\pm$ 0.006}\end{tabular} &  \begin{tabular}[c]{@{}c@{}} \vspace{-2mm} 0.984\\ {\tiny $\pm$ 0.000}\end{tabular} &  \begin{tabular}[c]{@{}c@{}} \vspace{-2mm} 0.972\\ {\tiny $\pm$ 0.002}\end{tabular} &  \begin{tabular}[c]{@{}c@{}} \vspace{-2mm} 0.871\\ {\tiny $\pm$ 0.007}\end{tabular} & \green \begin{tabular}[c]{@{}c@{}} \vspace{-2mm} 0.932\\ {\tiny $\pm$ 0.003}\end{tabular} &  \begin{tabular}[c]{@{}c@{}}\textbf{ \vspace{-2mm} 0.934}\\ {\tiny $\pm$ 0.005}\end{tabular} &  \begin{tabular}[c]{@{}c@{}} \vspace{-2mm} 0.829\\ {\tiny $\pm$ 0.008}\end{tabular}\\ \hline 
 zipper &  \begin{tabular}[c]{@{}c@{}} \vspace{-2mm} 0.889\\ {\tiny - } \end{tabular} & \green \begin{tabular}[c]{@{}c@{}}\textbf{ \vspace{-2mm} 0.968}\\ {\tiny $\pm$ 0.002}\end{tabular} &  \begin{tabular}[c]{@{}c@{}} \vspace{-2mm} 0.775\\ {\tiny $\pm$ 0.011}\end{tabular} &  \begin{tabular}[c]{@{}c@{}} \vspace{-2mm} 0.870\\ {\tiny $\pm$ 0.002}\end{tabular} &  \begin{tabular}[c]{@{}c@{}} \vspace{-2mm} 0.859\\ {\tiny $\pm$ 0.002}\end{tabular} &  \begin{tabular}[c]{@{}c@{}} \vspace{-2mm} 0.797\\ {\tiny $\pm$ 0.138}\end{tabular} & \green \begin{tabular}[c]{@{}c@{}}\textbf{ \vspace{-2mm} 0.972}\\ {\tiny $\pm$ 0.002}\end{tabular} &  \begin{tabular}[c]{@{}c@{}} \vspace{-2mm} 0.950\\ {\tiny $\pm$ 0.006}\end{tabular} &  \begin{tabular}[c]{@{}c@{}} \vspace{-2mm} 0.903\\ {\tiny $\pm$ 0.006}\end{tabular}\\ \hline 
\gray mean & \gray \begin{tabular}[c]{@{}c@{}} \vspace{-2mm} 0.908\\ {\tiny $\pm$ 0.088}\end{tabular} & \green \begin{tabular}[c]{@{}c@{}}\textbf{ \vspace{-2mm} 0.953}\\ {\tiny $\pm$ 0.068}\end{tabular} & \gray \begin{tabular}[c]{@{}c@{}} \vspace{-2mm} 0.823\\ {\tiny $\pm$ 0.120}\end{tabular} & \gray \begin{tabular}[c]{@{}c@{}} \vspace{-2mm} 0.934\\ {\tiny $\pm$ 0.045}\end{tabular} & \gray \begin{tabular}[c]{@{}c@{}} \vspace{-2mm} 0.925\\ {\tiny $\pm$ 0.043}\end{tabular} & \gray \begin{tabular}[c]{@{}c@{}} \vspace{-2mm} 0.775\\ {\tiny $\pm$ 0.159}\end{tabular} & \green \begin{tabular}[c]{@{}c@{}}\textbf{ \vspace{-2mm} 0.879}\\ {\tiny $\pm$ 0.108}\end{tabular} & \gray \begin{tabular}[c]{@{}c@{}} \vspace{-2mm} 0.843\\ {\tiny $\pm$ 0.144}\end{tabular} & \gray \begin{tabular}[c]{@{}c@{}} \vspace{-2mm} 0.814\\ {\tiny $\pm$ 0.116}\end{tabular}\\ \hline 

\end{tabular}
}
\end{center}
\label{table:backbone}

\end{table*}

In this experiment we study the performances of various ImageNet classifiers as feature extractors for FAVAE. Results are listed in \tabref{table:backbone}. The Vanilla VAE is our baseline model that targets only pixel reconstruction. We also added scores from \citet{Dehaene2020} who tested several anomaly localization algorithms. For each dataset we reported the best score reported in their work.

For VGG16 we model the input of the 2nd, 3rd and 4th max pooling (right before activation). For ResNet18 we model features from conv2x, conv3x and conv4x. For YOLOv3 we model a single layer of feature - the one with the greatest resolution. YOLOv3 is an interesting model since it has been trained with supervision to upsample and merge high semantic features with low semantic features. This could be an alternative to the handcrafted upsampling and sum routine that we use to merge layers when using other feature extractors. 

For anomaly localization (Pixel AUROC), we observe that the VGG16 version is better or similar on every dataset except \textit{tile}. 
FAVAE based on VGG16 improves the \textit{Ref} scores from \citet{Dehaene2020}, which involve 8 different models and a comparatively slow gradient-based method. On average the reference score (0.908 $\pm$ 0.088) and vanilla VAE score (0.823 $\pm$ 0.120) are improved  by VGG16 (0.953 $\pm$ 0.068).

For anomaly detection (Image AUROC) VGG16 is still the best solution on average but is superior to the other feature extractors on only 11 over 15 datasets. On average vanilla score (0.775 $\pm$ 0.159) is improved by VGG16 (0.879 $\pm$ 0.108).
On both detection and localization tasks the improvement over vanilla VAE and \textit{Ref} are significant.

\end{document}